\documentclass[a4paper,11pt]{article}

\usepackage{fourier}
\usepackage{color}
 \usepackage{graphicx}
\usepackage{url}
\usepackage[affil-it]{authblk}
\usepackage{amsmath}
\usepackage{wrapfig}

\usepackage[T1]{fontenc}
\usepackage{times}

 \usepackage[nameinlink]{cleveref}

\usepackage[figuresleft]{rotating}
\usepackage{adjustbox}

\usepackage{titlesec}
\usepackage[font=small,skip=5pt]{caption}

\DeclareMathOperator*{\argmin}{argmin}
\usepackage{rotating}

\makeatletter
\newcommand{\smallsym}[2]{#1{\mathpalette\make@small@sym{#2}}}
\newcommand{\make@small@sym}[2]{%
  \vcenter{\hbox{$\m@th\downgrade@style#1#2$}}%
}
\newcommand{\downgrade@style}[1]{%
  \ifx#1\displaystyle\scriptstyle\else
    \ifx#1\textstyle\scriptstyle\else
      \scriptscriptstyle
  \fi\fi
}
\makeatother

\titlespacing*\section{0pt}{7pt plus 2pt minus 2pt}{3pt plus 2pt minus 2pt}
\titlespacing*\subsection{0pt}{10pt plus 4pt minus 2pt}{3pt plus 2pt minus 2pt}

\begin{document}

\title{Patch based Colour Transfer using SIFT Flow}

\author{Hana Alghamdi \& Rozenn Dahyot 
}
\affil{School of Computer Science \& Statistics\\ Trinity College Dublin, Ireland}

\date{}
\maketitle
\thispagestyle{empty}

\begin{abstract}

We propose a new colour transfer method with Optimal Transport (OT) to transfer the colour of a source image to match the colour of a target image of the same scene that may exhibit large motion changes between images. By definition  OT does not take into account any available information about correspondences when computing the optimal solution. To tackle this problem we propose to encode overlapping neighborhoods of pixels using both their colour and spatial correspondences  estimated using motion estimation. We solve the high dimensional problem in 1D space using an iterative projection approach.  We further introduce  smoothing  as  part  of  the  iterative  algorithms  for  solving optimal transport  namely Iterative Distribution Transport (IDT) and its variant the Sliced Wasserstein Distance (SWD). Experiments show quantitative and qualitative improvements over previous state of the art colour transfer methods.

\end{abstract}
\textbf{Keywords:} Optimal Transport, Nadaraya-Watson estimator, Iterative Distribution Transfer, Sliced Wasserstein Distance, Colour Transfer

\section{Introduction}
\label{sec:intro}

Colour variations between photographs often happen due to illumination changes, using different cameras, different in-camera settings or due to tonal adjustments of the users. Colour transfer methods have been developed to transform a source colour image into a specified target colour image to match colour statistics or eliminate colour variations between different photographs. Colour transfer has many applications in image processing problems,  ranging from generating colour consistent image mosaicing and stitching \cite{brown2007automatic} to colour enhancement and style manipulation \cite{hwang2014color}.


When computing the transfer function, considering colour information only does not take into account the fact that coherent colours should be transferred to neighboring pixels, which can create results with blocky artifacts emphasizing JPEG compression blocks, or increase noise. To tackle this problem, Alghamdi  et al. \cite{Alghamdi2019} proposed the Patch based Colour Transfer (PCT\_OT) approach that encodes overlapping neighborhoods of pixels, taking into account both their colour and pixel positions. PCT\_OT algorithm shows improvement over the state of the art methods but also shows limitations by creating shadow artifacts when there are large changes between target and source images. In this paper we  propose to improve PCT\_OT  by first
improving the data preparation step for defining patches thanks to  SIFT flow   \cite{SIFTflow2010}. We estimate motions between images using  SIFT flow approach and incorporate the spatial correspondence information in the encoded overlapping neighborhoods of pixels. This formulation makes OT \textit{implicitly} take into account correspondences information when computing the optimal solution. Our second contribution is to introduce smoothing as part of the iterative algorithms for solving optimal transport  namely Iterative Distribution Transport (IDT) and its variant the Sliced Wasserstein Distance (SWD).

\section{PCT\_OT with SIFT Flow }

\subsection{ Combine colour and spatial information}\label{subsection: PCTOT_Combine_colour_position}
The spatial information for the target image is calculated using
SIFT flow method which estimates dense spatial correspondences by robustly aligning complex scene pairs containing significant spatial differences \cite{SIFTflow2010},  while in \textit{PCT\_OT} \cite{Alghamdi2019}  the original pixel positions in the grid coordinate of the image are used. Using correspondences  will allow colour transfer between images that contain moving objects and overcome the limitations in \textit{PCT\_OT}. More specifically, let $y^p$ be the 2D pixel position of the target image to be computed, and let $\mathbf{p}=(a,b)$ be the 2D grid coordinate of the target image and $\mathbf{w}(\mathbf{p})=(u(\mathbf{p}),v(\mathbf{p}))$ be the flow vector at $\mathbf{p}$ computed using SIFT flow method, then $y^p= \mathbf{p}+\mathbf{w}(\mathbf{p}) =(a + u(\mathbf{p}), b + v(\mathbf{p}))$ is the new pixel position in the target image that match a pixel position in the source image. The pixel's colour $y^c$  and its pixel position $y^p$ are concatenated  into a vector $y=(y^c, y^p)^T$ such that $\dim(y)=\dim(y^c)+\dim(y^p)$.  The source image keeps the grid coordinate of the image as pixel positions, i.e $x^p=\mathbf{p}$ and similarly to the target image the pixel's colour $x^c$  and its pixel position $x^p$ are concatenated  into a vector $x=(x^c, x^p)^T$ such that $\dim(x)=\dim(x^c)+\dim(x^p)$.

\subsection{Data normalisation}
\label{subsection: PCTOT_coord_sys}
Since the colours have integer values from 0 to 255, and the spatial values can be anything depending on the size of the image. In order to produce  consistent results regardless of the size of the image and better control parameters, we  normalize all the colour and position coordinates to lie between $0$ and $255$ to create a hypercube in $\mathbb{R}^d$. We then stretch that space in the direction of the spatial coordinates by a factor $w$  to make it harder to move the pixels in the spatial domain than in the colour domain, because since we are focusing on transferring colour between images of a same scene, we know that the scenes are overlapped and hence the more overlapped areas we have the higher $w$ value we can set. 

\subsection{Create patch vectors}\label{subsection: patch_vectors}
Similarly to \textit{PCT\_OT} \cite{Alghamdi2019} we encode overlapping  neighborhoods  of  pixels  to preserve local topology information. Starting from the origin of the coordinate system of the images (upper left corner),  we use a sliding window operation of window size $k \times k$ to extract overlapping patches. From each individual patch  we create a high dimensional  vector  in $\mathbb{R}^{d\times k \times k}$. We apply this process to the source and target images to create patch vector sets $\lbrace x_i \rbrace$ and $\lbrace y_j \rbrace$ for each respectively.

\section{Smoothed solution for 1D Optimal Transport}

The OT problem consists of estimating the  minimum cost (referred to as the Wasserstein Distance \cite{ villani2008optimal} or as the Earth Mover's Distance \cite{rubner2000earth}) of transfering a source distribution to a target distribution.  
As a byproduct of OT distance estimation, the mapping $\phi$ itself between the two distributions is also provided.
Monge's formulation of OT  \cite{villani2008optimal} defines the deterministic coupling $y=\phi(x)$ between random vectors $x\sim f(x)$  and $y \sim g(y)$ that capture the colour information of the source and target images respectively, and its solution minimizes the total transportation cost:
\begin{equation}
    \argmin_{\phi} \;  \int \; {\lVert {{x}-\phi({x})}\rVert}^2  \; f({x})\ dx \quad  \text{such that}: \quad f(x)=g(\phi(x))   \  | \det \nabla \phi(x)| 
\label{monge}
\end{equation}
with $f$ the probability density function (pdf) of $x$ and $g$ the pdf of $y$. The solution for $\phi$ can be found using  existing algorithms  such as linear programming, and the Hungarian and Auction algorithms \cite{santambrogio2015optimal}. However, in practice it is difficult to find a solution for colour images when  $\dim(x)=\dim(y)=d>1$  as the computational complexity of these solvers increases in multidimensional spaces \cite{villani2003topics}. But for $d=1$, with $x, y \in\mathbb{R}$,  a solution for $\phi$ is straightforward and can be defined using the increasing rearrangement \cite{villani2008optimal}:
\begin{equation}
     \phi^{OT}= G^{-1}\circ F \label{incRearrangement}
\end{equation}
where $F$  and $G$ are the cumulative distributions of the colour values in the source and target images respectively.

\subsection{Iterative Distribution Transfer (IDT)}

The 1D solution $\phi^{OT}$ Eq.  (\ref{incRearrangement}) has been used to tackle problems in multidimensional colour spaces and of particular interest is  the Iterative Distribution Transfer (IDT) algorithm for colour transfer proposed by  Piti\'{e} et al. \cite{Pitie_CVIU2007}. They proposed to iteratively project  colour values $\lbrace x_i \rbrace_{i=1}^n$ and $\lbrace y_j \rbrace_{j=1}^m$ originally in $\mathbb{R}^d$ to a 1D subspace and solve the OT using $\phi^{OT}$ Eq. (\ref{incRearrangement}) in this 1D subspace and then propagate the solution back to $\mathbb{R}^d$ space. This operation is repeated with different directions in 1D space until convergence.  This strategy was inspired by the idea of the Radon Transform \cite{Pitie_CVIU2007} which states the following proposition: if the target and source colour points are aligned in all possible 1D  projective  spaces, then matching is also achieved in $\mathbb{R}^d$ space. Note that the implementation of IDT approximates $F$ and $G$  using cumulative histograms
which can  be considered as  a form of quantile matching but with irregular quantile increments derived from the cumulative histograms of the source and target images - as source and target quantiles do not match exactly, interpolation can be used to compute solution   \cite{Pitie_CVIU2007}. 

\subsection{Sliced-Wasserstein Distance (SWD)}

The Sliced Wasserstein Distance (SWD) algorithm follows from the iterative projection approach of IDT
 but computes  the 1D solution $\phi^{OT}$ with quantile matching  instead of cumulative histogram matching \cite{rabin2011wasserstein,BonneelJMIV2015}. More specifically,
 SWD  sorts the $n$ 1D projections of the source and target images respectively to define quantiles with regular increments of size $\frac{1}{n}$ between 0 and 1 for both source and target distributions. 
 The SWD algorithm can be computed in $\mathcal{O}(n\log(n))$ operations using a fast sorting algorithm \cite{rabin2011wasserstein}. When a small number of observations are available, using SWD is best but with a large number of observations, histogram matching with IDT is more efficient.

\subsection{Smoothing $\phi^{OT}$ with Nadaraya Watson Estimator}

 Giving the correspondences $\{(x_i,y_i)\}_{i=1,\cdots,n}$, the Nadaraya Watson (NW) estimator is defined as follows:
\begin{equation}
\mathbb{E}[y|x] =\int y \ p(y|x)\  dy=  \int y\  \frac{ p(y,x)}{p(x)}\  dy  \simeq \frac{n^{-1}\sum_{i=1}^n y_{i} \ K_h(x-x_{i})}{n^{-1}\sum_{i=1}^n  K_h(x-x_{i})}=\phi^{NW}_h(x)
\end{equation} 
With this form NW  can be seen as locally weighted average of $\{ y_i\}_{i=1,\cdots,n}$, using a kernel as a weighting function where the bandwidth $h$ is the hyperparameter or scale parameter of the kernel, the larger the value of $h$ the more $\phi^{NW}_h$ gets smoother. 
We propose to smooth  $\phi^{OT}$ computed in IDT or SWD by using non-parametric Nadaraya Watson estimator. 
At each iteration $t$, following the step of calculating the optimal map $\phi^{OT}$, we feed the  OT estimated correspondences $\{(x_i,\phi^{OT}(x_i)) \}_{i=1}^n$ to NW estimator to compute a smoother OT solution, denoted as  $\phi^{OT}_h$,  defined as follows:
\begin{equation}
\phi^{OT}_h(x)=\frac{\sum_{i=1}^n \phi^{OT}(x_i) \ K_h(x-x_{i})}{\sum_{i=1}^n  K_h(x-x_{i})}
\label{NWeq}
\end{equation}

\begin{figure*}[!h]
\begin{center}
 \includegraphics[width = 1\linewidth, height =.27\linewidth]{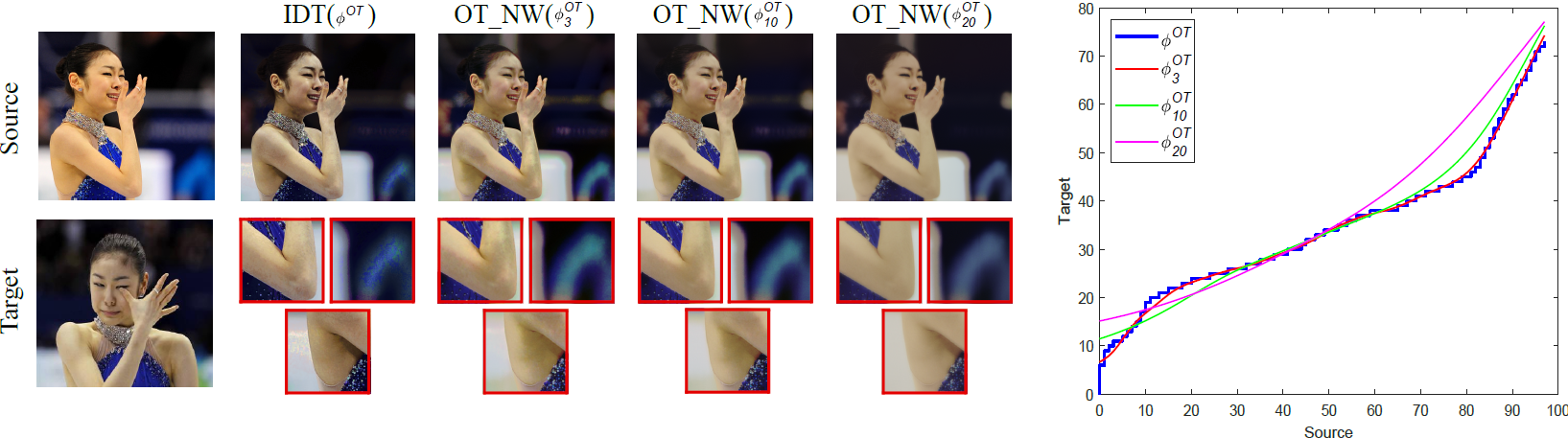}
\end{center}
   \caption{Results shows the smoothed Optimal Transport solution using non-parametric Nadaraya-Watson ($\phi^{OT}_h$) with different bandwidth values $h=\{3,10,20\}$.   Nadaraya-Watson significantly reduces the grainy artifacts produced by the original Optimal Transport mapping ($\phi^{OT}$), the bigger $h$ value the more smoothed mapping. The results processed without post processing step (best viewed in colour and zoomed in).}
\label{fig:OTNW_h}
\end{figure*}

Figure \ref{fig:OTNW_h} illustrates the effect of computing smoother OT solutions using NW with different bandwidth values on colour transfer compared with the original OT solution computed using IDT algorithm  \cite{Pitie_CVIU2007}. Optimal Transport solution is suitable in situations where the function that we need to estimate must satisfy important side conditions, such as being strictly increasing, and the non-parametric NW estimator on top of the OT solution can provide the smoothness required in the estimated function. In addition, one of the important characteristics of using OT and NW estimators is that they do not assume explicit expression controlled by parameters on the regression function  which makes them directly employable. In the following sections we are applying OT and NW smoothing in the relevant context of colour transfer where the the function that we need to estimate must satisfy the condition of being increasing function.


\section{Experimental Assessment}
 \label{sec:experiments}
We provide here quantitative and qualitative evaluations of our approach noted \texttt{OT\_NW} with comparisons to different state of the art colour transfer methods noted  \texttt{IDT} \cite{Pitie_CVIU2007}, \texttt{PMLS} \cite{hwang2014color}, \texttt{GPS/LCP} and \texttt{FGPS/LCP}  \cite{bellavia2018dissecting}, \texttt{L2} \cite{GroganCVIU19} and \texttt{PCT\_OT} \cite{Alghamdi2019}. In these evaluations we use image pairs with
similar content from an existing dataset provided by Hwang et al \cite{hwang2014color}. The dataset includes registered pairs of images (source and target) taken with different cameras and settings, and different illuminations and recolouring styles.

\subsection{Colour space and parameters settings}
We use  the RGB colour space where each pixel is represented by its 3D RGB colour values  and its 2D spatial position. Our patches with combined colour and spatial features create a vector in 125 dimensions ($5 \times 5\times 5$) for the RGB colours (3D) and position component (2D). We found patch size of $5 \times 5$ captures enough of a pixel's neighbourhood. 
We stretch the  hypercube space in $\mathbb{R}^d$ in the direction of the spatial coordinates by a factor $w=10$  to make it harder to move the pixels in the spatial domain than in the colour domain. We experimented with different bandwidth values and we found a fix value of $h=10$ gives best results.

\subsection{Evaluation metrics}
To quantitatively assess the recolouring results,  four metrics are used: peak signal to noise ratio (PSNR) \cite{salomon2004data}, structural similarity index (SSIM) \cite{wang2004image}, colour image difference (CID) \cite{preiss2014color} and feature similarity index (FSIMc) \cite{zhang2011fsim}. These metrics are often used when considering source and target images of the same content \cite{lissner2013image, oliveira2015probabilistic, hwang2014color, bellavia2018dissecting}. Note that the results using \texttt{PMLS} were provided by the authors \cite{hwang2014color}. It has already been shown in \cite{GroganCVIU19}  that \texttt{PMLS} performs better than  two other more recent techniques using correspondences \cite{park2016, Xia2017},  
so \texttt{PMLS} is the one reported here  with \cite{Alghamdi2019, GroganCVIU19} as algorithms that account for correspondences.


\subsection{Experimental Results}

\begin{figure}
\centering
\begin{tabular}{c}
\includegraphics[width=12.6cm]{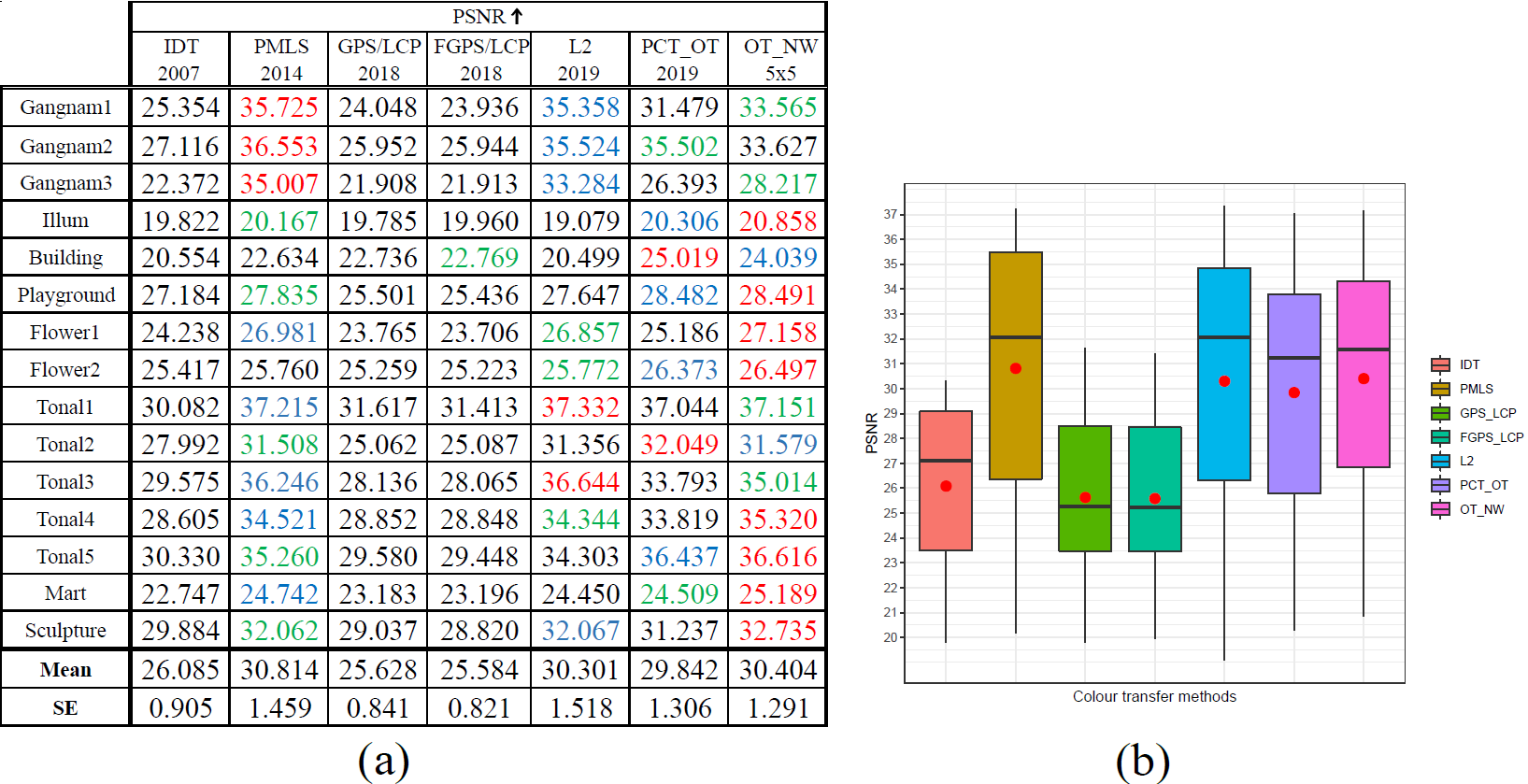}
\end{tabular}
\caption{Metric comparison, using PSNR \cite{salomon2004data}. (a) Red, blue, and green indicate $1^{st}$, $2^{nd}$, and $3^{rd}$ best performance respectively in the table (higher values are better), (b) visualized in box plot (best viewed in colour and zoomed in).}
\label{fig:psnr}
\end{figure}

\begin{figure}
\centering
\begin{tabular}{c}
\includegraphics[width=12.6cm]{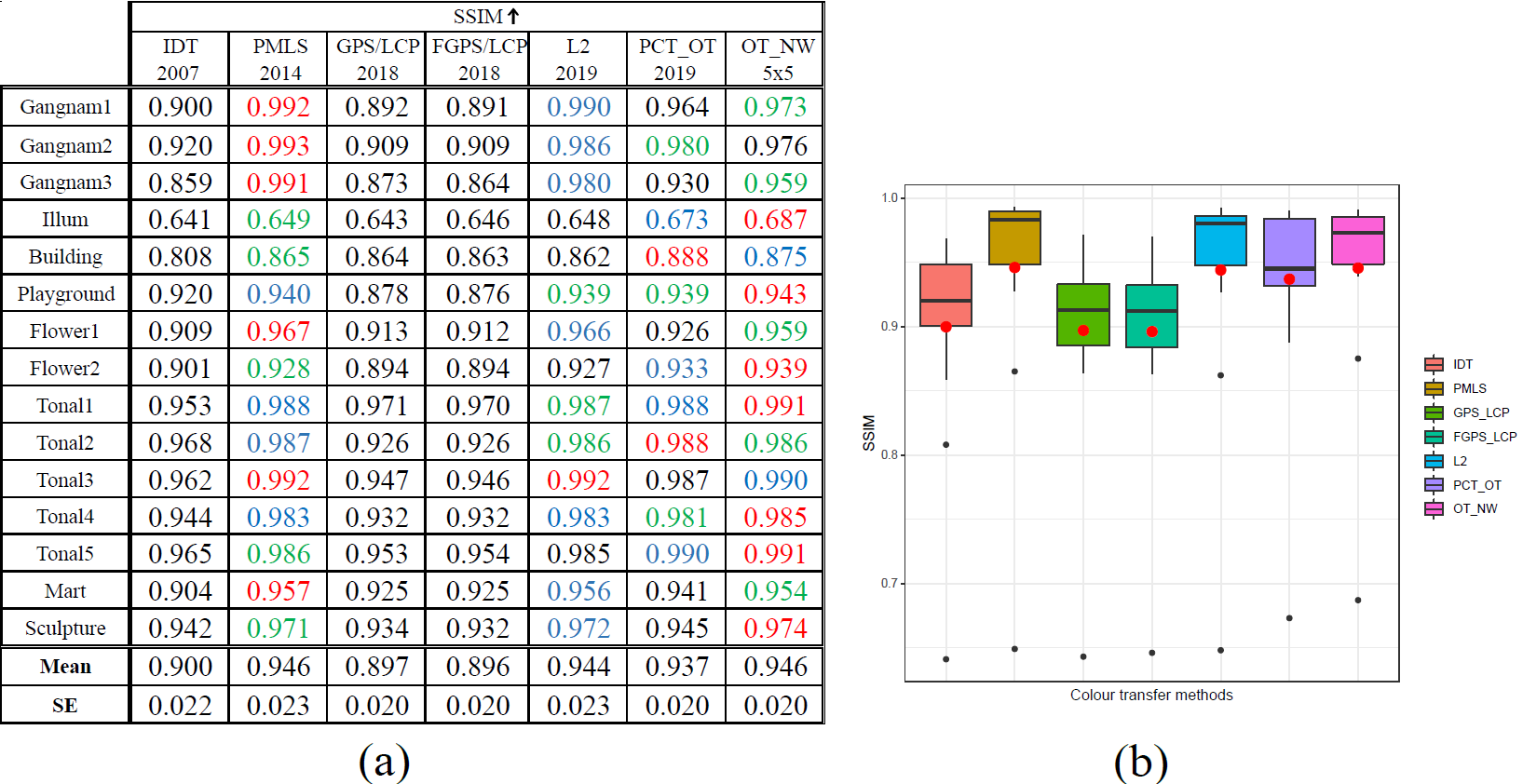}
\end{tabular}
\caption{Metric comparison, using SSIM \cite{wang2004image}. (a) Red, blue, and green indicate $1^{st}$, $2^{nd}$, and $3^{rd}$ best performance respectively in the table (higher values are better), (b) visualized in box plot (best viewed in colour and zoomed in).}
\label{fig:ssim}
\end{figure}

\begin{figure}
\centering
\begin{tabular}{c}
\includegraphics[width=12.6cm]{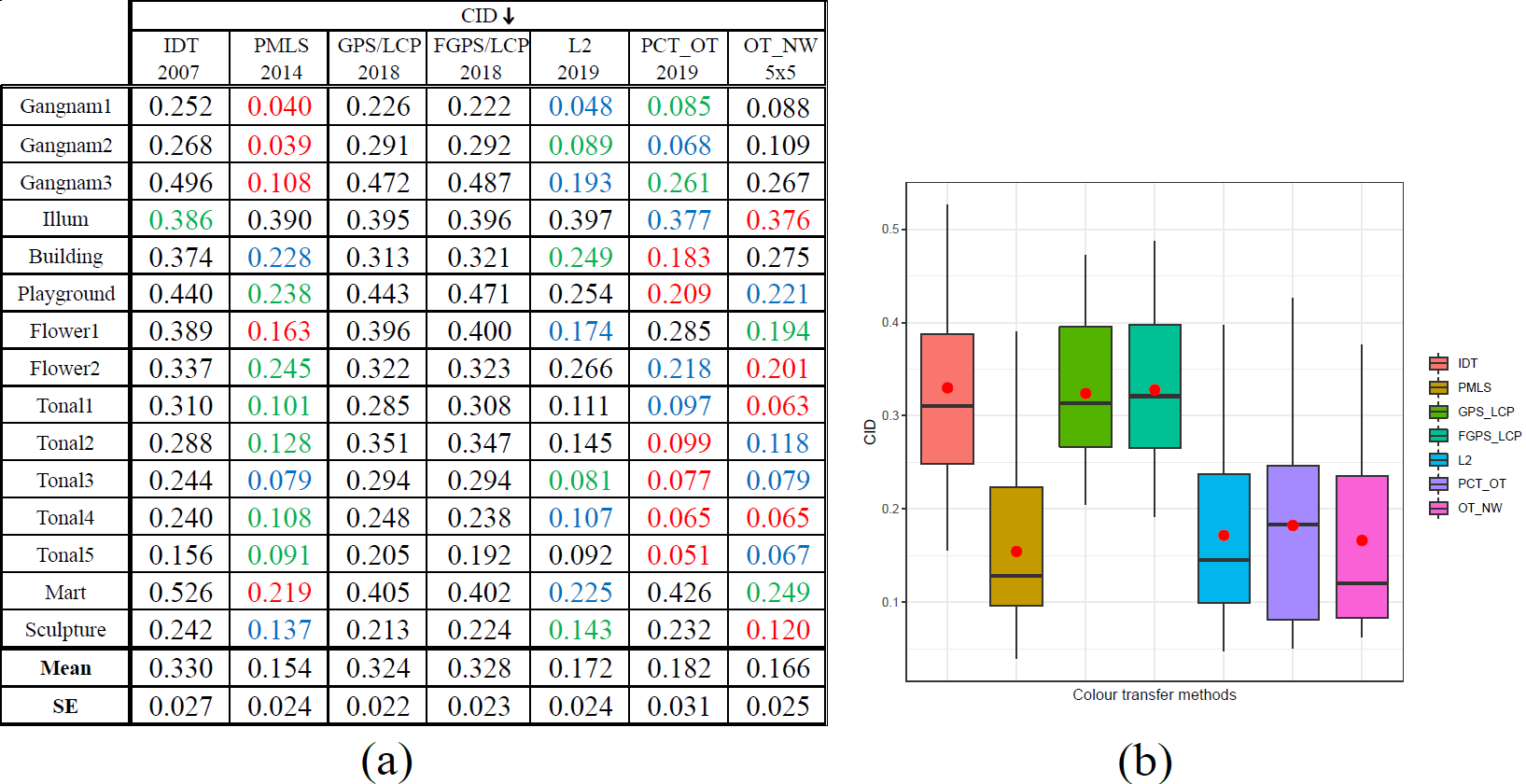}
\end{tabular}
\caption{Metric comparison, using CID \cite{preiss2014color}. (a) Red, blue, and green indicate $1^{st}$, $2^{nd}$, and $3^{rd}$ best performance respectively in the table (lower values are better), (b) visualized in box plot (best viewed in colour and zoomed in).}
\label{fig:cid}
\end{figure}

\begin{figure}
\centering
\begin{tabular}{c}
\includegraphics[width=12.6cm]{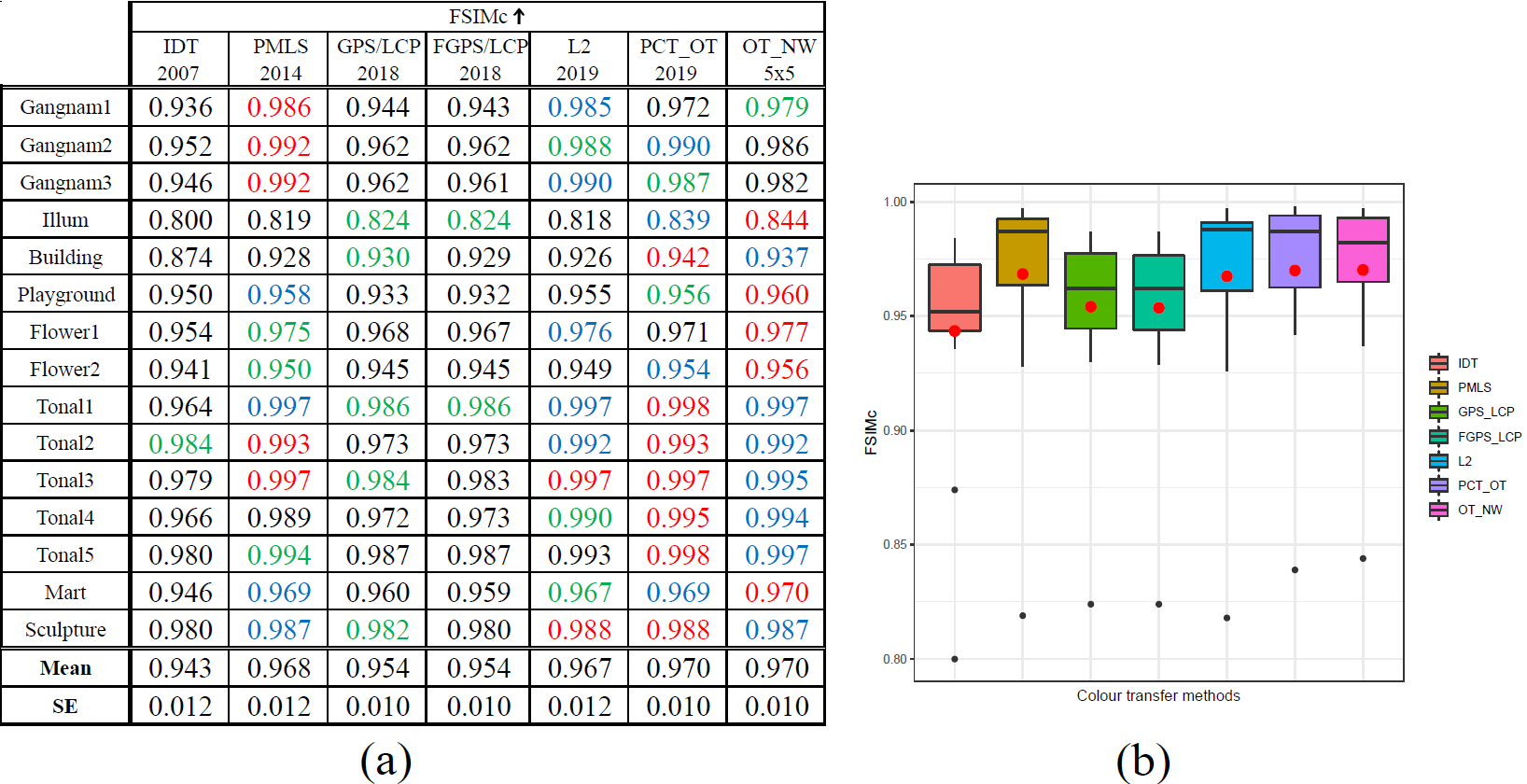}
\end{tabular}
\caption{Metric comparison, using FSIMc \cite{zhang2011fsim}. (a) Red, blue, and green indicate $1^{st}$, $2^{nd}$, and $3^{rd}$ best performance respectively in the table (higher values are better), (b) visualized in box plot (best viewed in colour and zoomed in).}
\label{fig:fsimc}
\end{figure}

\Cref{fig:psnr,fig:ssim,fig:cid,fig:fsimc} show detailed tables of quantitative results for each metric  alongside with box plots carrying  a lot of statistical details. The purpose of the  box plots is to visualize differences among methods and to show how close our method is to the state of the art algorithms. \Cref{fig:psnr} (b) and  \Cref{fig:fsimc} (b) shows PSNR and FSIMc metrics results respectively, by examining the box plots in both figures we see that the four methods \texttt{PMLS}, \texttt{L2}, \texttt{PCT\_OT} and \texttt{OT\_NW} are greatly overlap with each other, the median and mean values (the mean shown as red dots in the plots) are the highest among all algorithms and  are very close in value and the whiskers length almost similar indicating similar data variation and consistency.  \Cref{fig:ssim} (b) shows SSIM box plot, we can see that   \texttt{OT\_NW} performs similarly with   \texttt{PMLS} and \texttt{L2}  scoring highest values while here the median line of   \texttt{PCT\_OT}  box lies outside the three top algorithms scoring the lowest value among them. With CID metric in \Cref{fig:cid},   \texttt{OT\_NW} performs similarly with   \texttt{PMLS}, \texttt{L2} and \texttt{PCT\_OT}. In conclusion, the quantitative metrics show that our algorithm with Nadaraya Watson \texttt{OT\_NW} performs similarly with top methods \texttt{PMLS}, \texttt{L2} and \texttt{PCT\_OT} and outperforms the rest of  the state of the art algorithms.

Figure  \ref{fig:qualitative_strips}  provides qualitative results. For  clarity, the results are presented in image mosaics, created by switching between the target image and the transformed source image column wise (Figure \ref{fig:qualitative_strips}, top row). If the colour transfer is accurate, the resulting
mosaic should look like a single image (ignoring the small motion displacement between source and target images), otherwise column differences appear. As can be noted, our approach \texttt{OT\_NW} with Nadaraya Watson step is visually the best at removing the column differences.

\begin{figure}
\centering
\begin{tabular}{c}
\includegraphics[width = .93\linewidth, height =.4\linewidth]{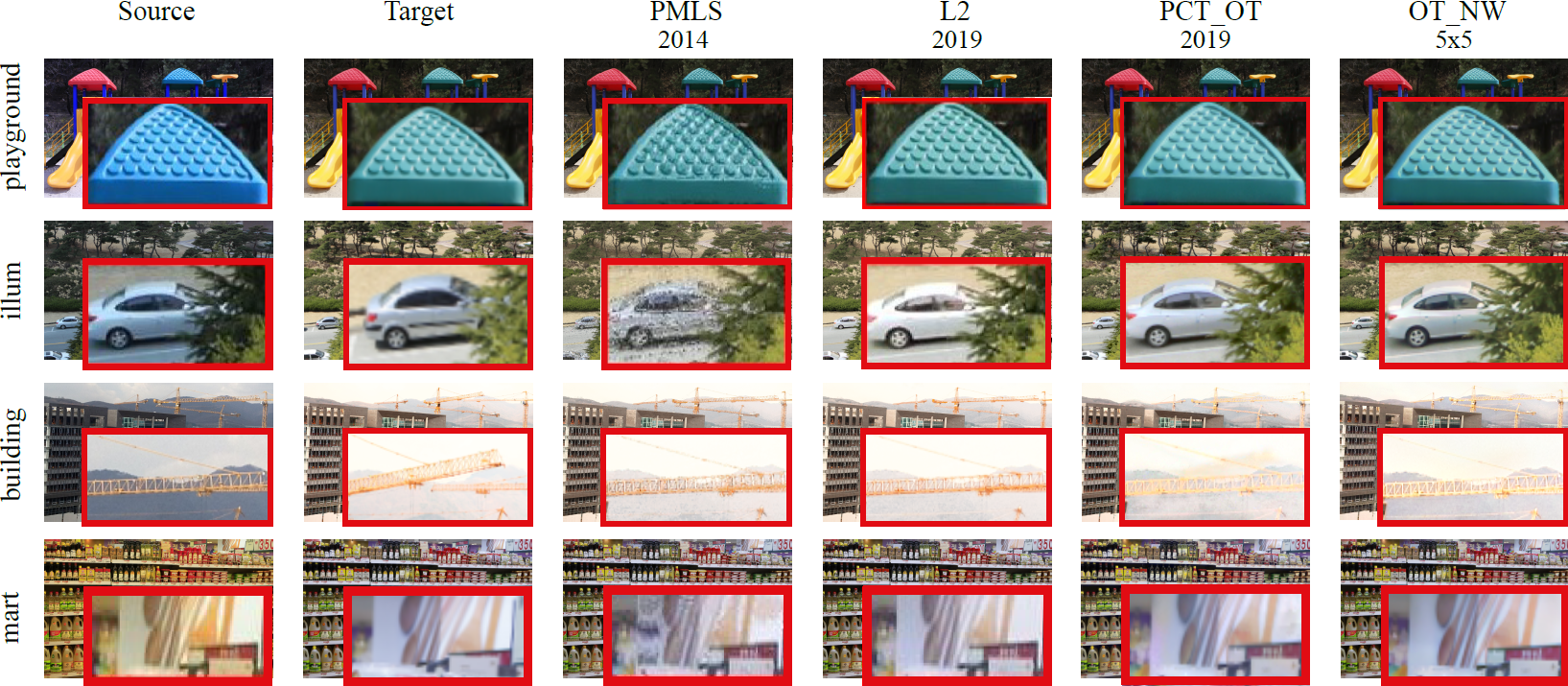}
\end{tabular}
\caption{A close up look at some of the results generated using the PMLS, L2, \textit{PCT\_OT} and our algorithm \textit{OT\_NW} (best viewed in colour and zoomed in).}
\label{fig:artifacts}
  
\end{figure}

While \texttt{PMLS} and \texttt{PCT\_OT} provide equivalent results to our method in terms of metrics measures, \texttt{PMLS} on the one hand introduces visual artifacts if the input images are not registered correctly (Figure \ref{fig:artifacts}), while our method is robust to registration errors. Note that although the accuracy of the PSNR, SSIM, CID and FSIMc metrics relies on the fact that the input images are registered correctly; if this is not the case, these metrics may not accurately capture all artifacts (Figures  \ref{fig:qualitative_strips} and \ref{fig:artifacts}). In addition, due to the Nadaraya Watson smoothing step in our algorithm,  our approach allows us to create a smoother colour transfer result, and can also alleviate JPEG compression artifacts and noise (cf. Figure \ref{fig:artifacts} for comparison). On the other hand, \texttt{PCT\_OT} can create shadow artifacts when there are large changes between target and source images (Figure \ref{fig:artifacts}, in example `building'), while our method \texttt{OT\_NW} can correctly transfer colours between images that contain significant spatial differences and alleviates the shadow artifacts, as can be seen in Figure \ref{fig:artifacts} with examples 'illum', 'mart' and 'building'.

\section{Conclusion}

Several contributions to colour transfer with OT have been made in this paper, showing quantitative and qualitative improvements over state of the art methods. In particular, first, correspondences information as well as colour content of pixels are both encoded in the high dimensional feature vectors, and second, we introduced  smoothing  as  part  of  the  iterative  algorithms  for  solving optimal transport  namely Iterative Distribution Transport (IDT) and its variant the Sliced Wasserstein Distance (SWD). The algorithm allows denoising, artifact removal as well as smooth colour transfer  between images that may contain large motion changes.


\begin{center}
    \begin{sideways}
         \begin{minipage}{1.3\linewidth}
                    \includegraphics[width=\linewidth,keepaspectratio]{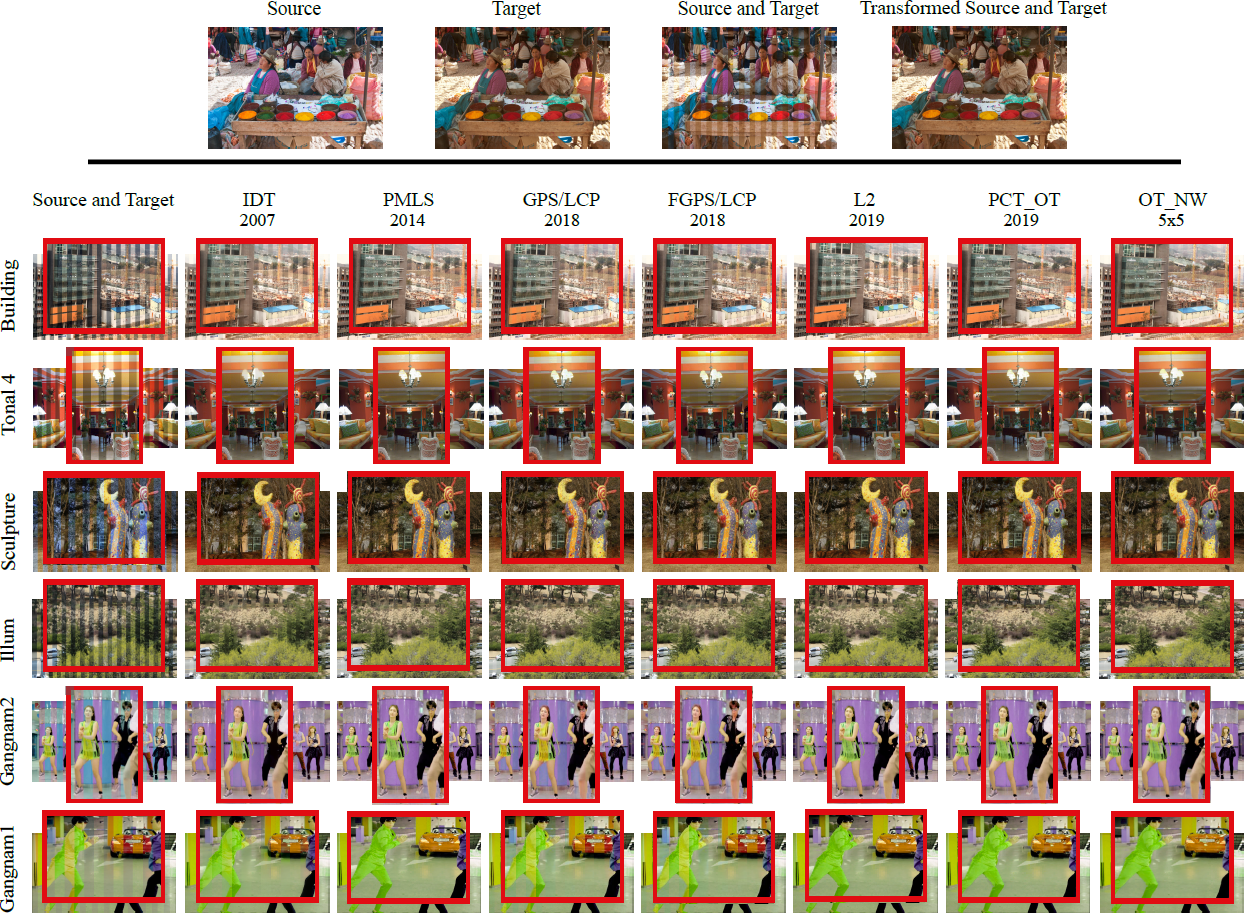}
                    \captionof{figure}{A close up look at some of the results generated using the PMLS, L2, \textit{PCT\_OT} and our algorithm \textit{OT\_NW} (best viewed in colour and zoomed in).}
         \vspace{2cm}
         \label{fig:qualitative_strips}
         \end{minipage}
    \end{sideways}
    \end{center}

\section*{Acknowledgments}
This work is partly funded by a scholarship from Umm Al-Qura University, Saudi Arabia, and in part by a research grant from Science Foundation Ireland (SFI) under the Grant Number 15/RP/2776, and the ADAPT Centre for Digital Content Technology (www.adaptcentre.ie) that is funded under the SFI Research Centres Programme (Grant 13/RC/2106) and is co-funded under the European Regional Development Fund.

\bibliographystyle{unsrt}  
{\small \bibliography{references} } 

\begin{thebibliography}{10}

\bibitem{brown2007automatic}
M.~Brown and D.~G. Lowe.
\newblock Automatic panoramic image stitching using invariant features.
\newblock {\em International Journal of Computer Vision}, 74(1):59--73, Aug
  2007.

\bibitem{hwang2014color}
Y.~Hwang, J.~Lee, I.~S. Kweon, and S.~J. Kim.
\newblock Color transfer using probabilistic moving least squares.
\newblock In {\em IEEE Conf. on Computer Vision and Pattern Recognition
  (CVPR)}, pages 3342--3349, June 2014.

\bibitem{Alghamdi2019}
H.~{Alghamdi}, M.~{Grogan}, and R.~{Dahyot}.
\newblock Patch-based colour transfer with optimal transport.
\newblock In {\em 2019 27th European Signal Processing Conference (EUSIPCO)},
  pages 1--5, Sep. 2019.
\newblock \url{https://github.com/leshep/PCT_OT}.

\bibitem{SIFTflow2010}
C.~{Liu}, J.~{Yuen}, and A.~{Torralba}.
\newblock Sift flow: Dense correspondence across scenes and its applications.
\newblock {\em IEEE Transactions on Pattern Analysis and Machine Intelligence},
  33(5):978--994, May 2011.
\newblock \url{https://people.csail.mit.edu/celiu/SIFTflow/}.

\bibitem{villani2008optimal}
C.~Villani.
\newblock {\em Optimal transport: old and new}, volume 338.
\newblock Springer Science \& Business Media, 2008.

\bibitem{rubner2000earth}
Y.~Rubner, C.~Tomasi, and L.~J. Guibas.
\newblock The earth mover's distance as a metric for image retrieval.
\newblock {\em International Journal of Computer Vision}, 40(2):99--121, Nov
  2000.

\bibitem{santambrogio2015optimal}
F.~Santambrogio.
\newblock Optimal transport for applied mathematicians.
\newblock {\em Birk{\"a}user, NY}, pages 99--102, 2015.

\bibitem{villani2003topics}
C.~Villani.
\newblock {\em Topics in optimal transportation}.
\newblock Number~58. American Mathematical Soc., 2003.

\bibitem{Pitie_CVIU2007}
F.~Piti\'e, A.~C. Kokaram, and R.~Dahyot.
\newblock Automated colour grading using colour distribution transfer.
\newblock {\em Computer Vision and Image Understanding}, 107(1):123 -- 137,
  2007.
\newblock \url{https://github.com/frcs/colour-transfer}.

\bibitem{rabin2011wasserstein}
J.~Rabin, G.~Peyr{\'e}, J.~Delon, and M.~Bernot.
\newblock Wasserstein barycenter and its application to texture mixing.
\newblock In {\em Scale Space and Variational Methods in Computer Vision},
  pages 435--446. Springer Berlin Heidelberg, 2012.

\bibitem{BonneelJMIV2015}
N.~Bonneel, J.~Rabin, G.~Peyr{\'e}, and H.~Pfister.
\newblock Sliced and radon wasserstein barycenters of measures.
\newblock {\em Journal of Mathematical Imaging and Vision}, 51(1):22--45, Jan
  2015.

\bibitem{bellavia2018dissecting}
F.~Bellavia and C.~Colombo.
\newblock Dissecting and reassembling color correction algorithms for image
  stitching.
\newblock 27(2):735--748, Feb 2018.

\bibitem{GroganCVIU19}
M.~Grogan and R.~Dahyot.
\newblock L2 divergence for robust colour transfer.
\newblock {\em Computer Vision and Image Understanding}, 181:39--49, 2019.
\newblock \url{https://github.com/groganma/gmm-colour-transfer}.

\bibitem{salomon2004data}
D.~Salomon.
\newblock {\em Data compression: the complete reference}.
\newblock Springer Science \& Business Media, 2004.

\bibitem{wang2004image}
Z.~Wang, A.~C. Bovik, H.~R. Sheikh, and E.~P. Simoncelli.
\newblock Image quality assessment: from error visibility to structural
  similarity.
\newblock 13(4):600--612, April 2004.

\bibitem{preiss2014color}
J.~Preiss, F.~Fernandes, and P.~Urban.
\newblock Color-image quality assessment: From prediction to optimization.
\newblock 23(3):1366--1378, March 2014.

\bibitem{zhang2011fsim}
L.~Zhang, L.~Zhang, X.~Mou, and D.~Zhang.
\newblock Fsim: A feature similarity index for image quality assessment.
\newblock 20(8):2378--2386, Aug 2011.

\bibitem{lissner2013image}
I.~Lissner, J.~Preiss, P.~Urban, M.~S. Lichtenauer, and P.~Zolliker.
\newblock Image-difference prediction: From grayscale to color.
\newblock 22(2):435--446, Feb 2013.

\bibitem{oliveira2015probabilistic}
M.~Oliveira, A.~D. Sappa, and V.~Santos.
\newblock A probabilistic approach for color correction in image mosaicking
  applications.
\newblock 24(2):508--523, Feb 2015.

\bibitem{park2016}
J.~Park, Y.~Tai, S.~N. Sinha, and I.~S. Kweon.
\newblock Efficient and robust color consistency for community photo
  collections.
\newblock In {\em IEEE Conf. on Computer Vision and Pattern Recognition
  (CVPR)}, pages 430--438, June 2016.

\bibitem{Xia2017}
M.~Xia, J.~Y. Renping, X.~M. Zhang, and J.~Xiao.
\newblock Color consistency correction based on remapping optimization for
  image stitching.
\newblock In {\em IEEE Int. Conf. on Computer Vision Workshops (ICCVW)}, pages
  2977--2984, Oct 2017.

\end{thebibliography}

\end{document}